\pdfoutput=1
\documentclass[11pt]{article}

\usepackage[final]{acl}

\usepackage{times}
\usepackage{latexsym}

\usepackage[T1]{fontenc}
\usepackage[utf8]{inputenc}

\usepackage{microtype}

\usepackage{inconsolata}

\usepackage{amsmath}
\usepackage{mathtools}
\usepackage{amssymb}
\usepackage{amsthm}
\usepackage{tabularx}
\usepackage{multirow}
\usepackage{multicol}
\usepackage{booktabs}
\usepackage{siunitx}
\usepackage{adjustbox}
\usepackage{{makecell}}
\setcellgapes{3pt}\makegapedcells

\usepackage{array,collcell}
\newcommand\AddLabel[1]{%
  \refstepcounter{equation}% increment equation counter
  (\theequation)% print equation number
  \label{#1}% give the equation a \label
}
\newcolumntype{M}{>{\hfil$\displaystyle}X<{$\hfil}}
\newcolumntype{L}{>{\collectcell\AddLabel}r<{\endcollectcell}}

\usepackage{graphicx}
\graphicspath{ {./images/} }
\usepackage[dvips]{epsfig}
\usepackage{svg}
\usepackage{relsize}
\usepackage{subcaption}
\usepackage{kotex}
\newcommand{\meanstd}[2]{#1_{\pm{#2}}}
\theoremstyle{plain}

\theoremstyle{definition}
\newtheorem{definition}{Definition}
\usepackage{algorithm}
\usepackage[noend]{algpseudocode}
\algnewcommand\algorithmicforeach{\textbf{for each}}
\algdef{S}[FOR]{ForEach}[1]{\algorithmicforeach\ #1\ \algorithmicdo}
\usepackage{tikz}
\usepackage{tikz-qtree}
\usepackage[linguistics]{forest}

\title{Probability Distribution Collapse:
A Critical Bottleneck to\\  Compact Unsupervised Neural Grammar Induction
}

\author{Jinwook Park \and Kangil Kim\thanks{Corresponding author.} \\
  AI Graduate School \\
  Gwangju Institute of Science and Technology\\
  \texttt{jinwookpark@gm.gist.ac.kr, kangil.kim.01@gmail.com}\\
}

\begin{document}
\maketitle

\begin{abstract}
Unsupervised neural grammar induction aims to learn interpretable hierarchical structures from language data. However, existing models face an expressiveness bottleneck, often resulting in unnecessarily large yet underperforming grammars. We identify a core issue, \textit{probability distribution collapse}, as the underlying cause of this limitation. We analyze when and how the collapse emerges across key components of neural parameterization and introduce a targeted solution, \textit{collapse-relaxing neural parameterization}, to mitigate it. Our approach substantially improves parsing performance while enabling the use of significantly more compact grammars across a wide range of languages, as demonstrated through extensive empirical analysis.
\end{abstract}

\section{Introduction}

% \paragraph{Background}
% Probabilistic Context-Free Grammars는 구조 정보를 해석하고 저장한다는 이점으로 널리 사용된다.
Formal grammars, such as context-free grammars, represent the structure of languages by formalizing the hierarchical organization of components of natural languages into human-understandable rules.
These rules enable logical interpretation of top-down or inclusion relationships within languages and support the control and use of hierarchical structures based on them.
This property has a potential to enable neural models to understand the structures, and there have been steady efforts to utilize grammars in natural language processing and computer vision area~\cite{Zhao2020VisuallyGC,Hong2021VLGrammarGG,Williams2023StructuredGM} via inducing high-quality grammars~\cite{shen2019ordered, dyer-etal-2016-recurrent, kim-etal-2019-unsupervised, wang2019tree, shen2021structformer, drozdov2019unsupervised, drozdov-etal-2020-unsupervised, kim2019compound}. In particular, Neural PCFGs (N-PCFGs)~\cite{kim2019compound} have successfully learned probabilistic context-free grammars (PCFGs) in an unsupervised setting by leveraging neural parameterization to estimate rule probability distributions from vector representations of symbols.

% \paragraph{Key Question}
Among various studies based on N-PCFGs~\cite{yang2021pcfgs,yang2022dynamic}, some have achieved higher performance by enhancing the expressive power of N-PCFGs. This improvement is grounded in the findings of conventional studies that did not utilize neural networks~\cite{petrov2006learning}, which demonstrated high parsing performance through an increased number of symbols, as well as in experimental results indicating that a greater number of grammar parameters can improve model performance~\cite{Buhai2019EmpiricalSO}.

%\paragraph{Gap}
However, before increasing grammar capacity for enhancing performance, can we be certain that current neural network architectures are truly designed to fully utilize the given capacity? 
This question is critical for interpretability, which requires both compact and accurate grammars. Yet, to date, the literature lacks analytical studies that directly address this important issue.

% 내용 줄이기
% \paragraph{Contribution}
We introduce \textit{Probability Distribution Collapse} (PDC) as a key bottleneck in constructing compact grammars. PDC implies indistinguishable probability distributions across many symbols, limiting grammar expressiveness even with a large number of symbols. We analyze this phenomenon in the neural parameterization process that maps symbol embeddings to rule probabilities, along with its training dynamics, and identify three underlying causes: 1) small dimension of embeddings and shallow neural network layers, 2) entangled scales of children embeddings in generating the probability, and 3) gradient explosion and dying ReLU collapsing symbol representations and corresponding probability projection. 
To this end, we propose \textit{collapse-relaxing neural parameterization} (CRNP) equipped to a recent work, N-PCFGs with parse-focusing~\cite{park-kim-2024-structural}, to address the causes via simple and comprehensive solutions.
In empirical validation on constituent parsing, our approach improves the upper bound of accuracy while maintaining the same grammar size under structural supervision, demonstrating its effectiveness without interference from the implicit behaviors of unsupervised neural grammar induction (UNGI).
Furthermore, in UNGI comparisons, our approach achieves strong performance even with highly compact grammars across multilingual benchmarks, including Penn TreeBank (PTB, English), Chinese TreeBank (CTB, Chinese), and SPMRL (Basque, French, German, Hebrew, Hungarian, Korean, Polish, Swedish).
Our code is publicly available at \url{https://github.com/GIST-IRR/CRNP}.

Our contributions are summarized as follows:
\begin{itemize}
    \item We introduce \textit{probability distribution collapse} as a bottleneck to inducing more compact grammars in an unsupervised setting.

    \item We investigate its causes within neural parameterization and propose a simple, yet effective solution, termed \textit{collapse-relaxing neural parameterization}.

    \item We provide extensive validation of our approach, demonstrating improvements in both upper bound and practical accuracy, as well as grammar compactness, on constituent parsing tasks for English and multilingual benchmarks.
\end{itemize}

\section{Background}

\paragraph{Notations of Probabilistic Context-Free Grammar}
\label{sec:pcfgs}
% PCFG에 대한 notation을 어떻게 정의하는가?
In this paper, we use the following notation for a PCFG $G=(S, N, P, \Sigma, R, \Pi)$: the root symbol $S$, finite sets of nonterminals $N$, preterminals $P$, terminal words $\Sigma$, production rules $R$, and production rule probabilities $\Pi$. $R$ consists of three types of rules:
\begin{equation*}
    \begin{array}{ll}
        S \rightarrow A
        &\text{where } A\in N \\
        A \rightarrow B\;C
        &\text{where }B, C\in (N\cup P) \\
        T \rightarrow \omega
        & \text{where }T\in P\text{ and } \omega\in\Sigma
    \end{array}
\end{equation*}

Note that the left-hand side of the rules is distinguished between nonterminal and preterminal categories. In this paper, we refer to the symbol on left-hand side as the parent and the symbols on right-hand side as the children.

\paragraph{Neural Parameterization}
% Neural Parameterization이란 무엇인가?
% Neural Parameterization means that parameterize probabilities of production rules from symbol embeddings by using neural network~\cite{kim2019compound}.
Neural parameterization is the process of using a neural network to parameterize the probabilities of production rules from symbol embeddings~\cite{kim2019compound}.
% In this paper, we call the neural network as neural parameterizer. 
N-PCFGs parametrize root, nonterminal, and preterminal symbols with independent different neural networks. In more detail, the symbol embeddings used in parameterizations are not shared.
For example, preterminal symbol embeddings in binary rule parameterizer and unary rule parameterizer are independent.
\begin{align}
    \pi_{S\rightarrow A}&=\frac{\exp(\mathbf{u}^{\mathsf{T}}_{A}f_{1}(\mathbf{w}_{S}))}{\sum_{A'\in\mathcal{N}}\exp(\mathbf{u}^{\mathsf{T}}_{A'}f_{1}(\mathbf{w}_{S}))},\label{eq:root_param}\\
    \pi_{A\rightarrow BC}&=\frac{\exp(\mathbf{u}^{\mathsf{T}}_{BC}\mathbf{w}_{A})}{\sum_{B'C'\in\mathcal{M}}\exp(\mathbf{u}^{\mathsf{T}}_{B'C'}\mathbf{w}_{A})},\label{eq:binary_param}\\
    \pi_{T\rightarrow \omega}&=\frac{\exp(\mathbf{u}^{\mathsf{T}}_{\omega}f_{2}(\mathbf{w}_{T}))}{\sum_{\omega'\in\Sigma}\exp(\mathbf{u}^{\mathsf{T}}_{\omega'}f_{2}(\mathbf{w}_{T}))}\label{eq:unary_param}
\end{align}
Where, $\mathbf{w}_{S}, \mathbf{w}_{A},\mathbf{w}_{T}$ represent root, nonterminal, preterminal symbol embeddings each, $f_1,f_2$ represent multi-layer perceptron composed by residual blocks, $\mathbf{u}^{\mathsf{T}}_{A},\mathbf{u}^{\mathsf{T}}_{BC},\mathbf{u}^{\mathsf{T}}_{\omega}$ represent weight matrix of each linear layer.
% 어떻게 학습하는지가 왜 필요하지?
The training objective of neural grammar induction is the maximization of sentence probabilities, and sentence probability is the sum of whole probabilities of derivable parse tree for given sentence.
The parse tree probabilities are calculated by inside algorithm using probability distribution obtained by neural parameterization.

\section{Probability Distribution Collapse}
\label{sec:pdc}

% 문제점에 대한 가설
\paragraph{Implicit Bottleneck of Expressiveness}
% Context
Probabilistic grammars have been central in computational linguistics, offering tractable analyses of capacity utilization.
In contrast, neural parameterizations lack such clarity due to the flexible and opaque mappings from symbol embeddings to probabilities.
As a result, it is challenging to validate that the neural probabilistic grammars utilize its full representational capacity.
% Limitations
Therefore, recent neural grammar induction methods frequently adopt overparameterized grammars to improve performance~\cite{yang2021pcfgs,yang2022dynamic,liu_simple_2023}, despite to the empirical availability of more compact grammars demonstrated in probabilistic learning frameworks~\cite{Stolcke_1994}.
This discrepancy suggests that neural parameterization imposes implicit constraints on grammatical expressiveness, limiting the effective use of available capacity.

\paragraph{Probability Distribution Collapse}

% PDC는 bottleneck을 유발하는 제약이다.
We propose \textit{Probability Distribution Collapse} (PDC) as the reason of the bottleneck.
PDC refers to a phenomenon where embeddings for distinct symbols are mapped to similar probability distributions, resulting in a large portion of dominating rules being shared.
We formalize PDC by using Jensen-Shannon Divergence (JSD) that measure the similarity between two probability distributions, where higher values indicate greater dissimilarity (See Appendix~\ref{sec:measures}).
Therefore, we define PDC as:

\begin{definition}
    Let $G$ be a neural grammar. Let $z_{i,G}$ and $z_{j,G}$ be distinct symbol embeddings in $G$ such that $z_{i,G}\neq z_{j,G}$. Let $f$ be a function that maps each symbol embedding to a probability distribution, yielding $p_i=f(z_{i,G})$ and $p_j=f(z_{j,G})$. We say that \textit{probability distribution collapse} occurs when the Jensen-Shannon Divergence between these distributions converges to zero, i.e., $\text{JSD}(p_i\|p_j)\rightarrow 0$.
    where the JSD is defined as
    \begin{equation*}
    \text{JSD}(P\|Q)=\frac{\text{KL}(P\|\frac{P+Q}{2})+\text{KL}(Q\|\frac{P+Q}{2})}{2}
    \end{equation*}
\end{definition}

% NP의 bias에 의해 PDC가 유발되면 표현력에 비해 다양성이 떨어진다.
% 다양성이 떨어지면 성능에 영향이 있다.
If a neural parameterization has an implicit bias that causes such PDC, a portion of symbols is inevitably wasted, and the grammar cannot represent the more diverse distributions available within its full capacity. 
This can degrade performance when structural diversity is required and limit even scalability in extending to more complex, real-world problems.

\paragraph{Empirical Evidence in N-PCFG}
\begin{figure}[t]
    \centering
    \begin{subfigure}[b]{0.8\columnwidth}
        \includegraphics[
            % inkscapelatex=false,
            width=5cm,
        ]{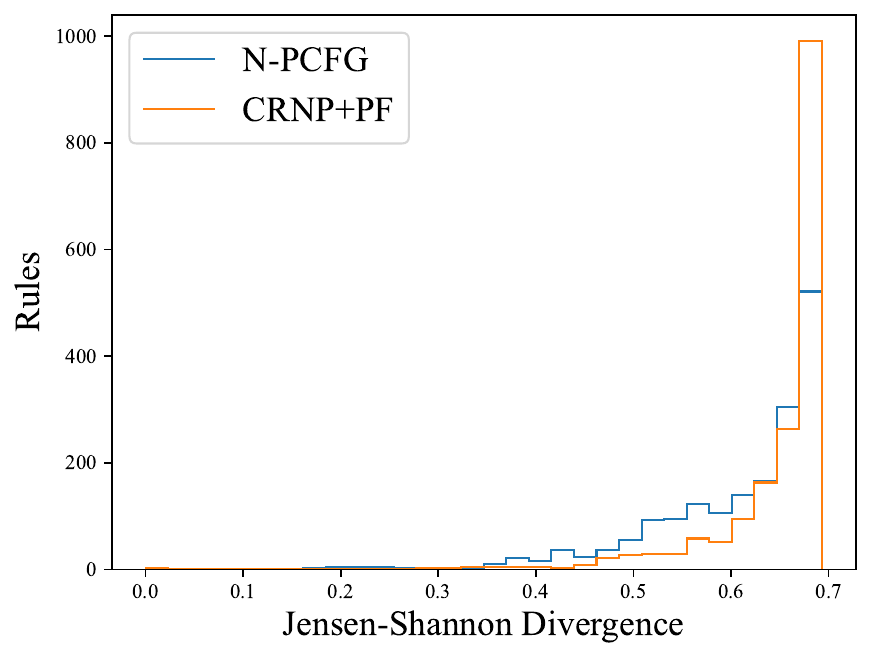}
        \caption{Dissimilarity of rule probability distributions}
        \label{fig:util_fgg}
    \end{subfigure}
    \hfill
    \begin{subfigure}[b]{0.48\columnwidth}
        \includegraphics[
            % inkscapelatex=false,
            width=\columnwidth,
        ]{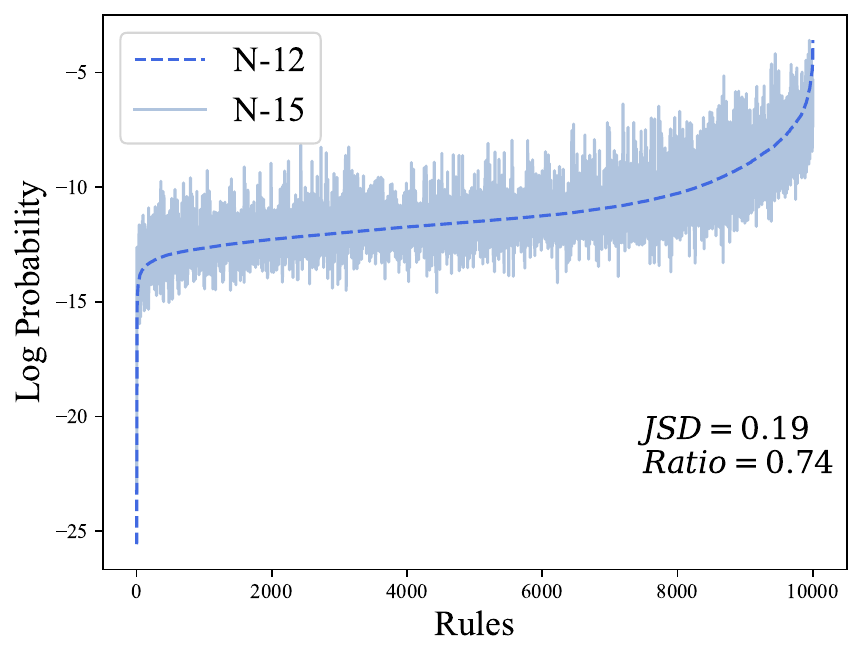}
        \caption{JSD = 0.19 (similar)}
        \label{fig:jsd_b}
    \end{subfigure}
    \begin{subfigure}[b]{0.48\columnwidth}
        \includegraphics[
            % inkscapelatex=false,
            width=\columnwidth,
        ]{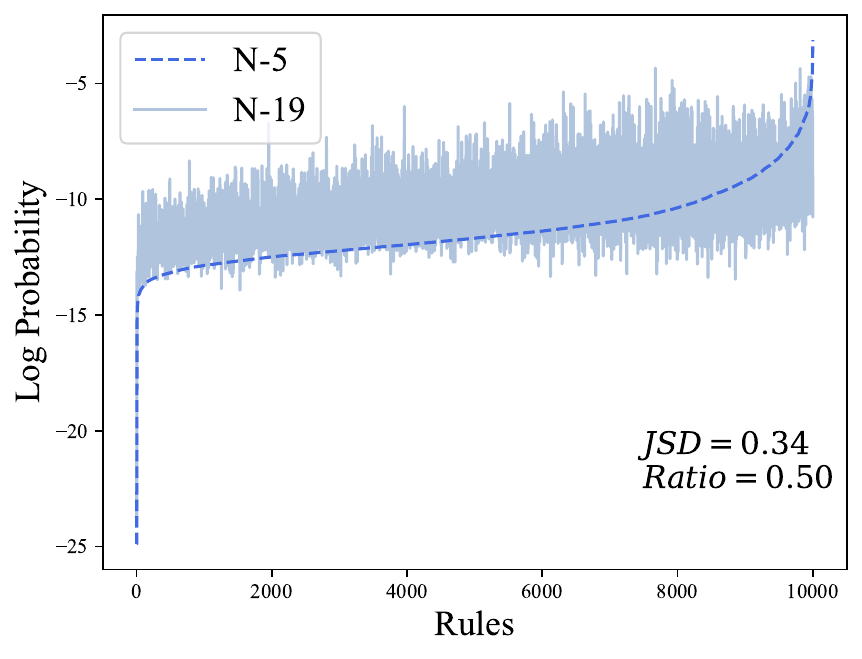}
        \caption{JSD = 0.34}
        \label{fig:jsd_c}
    \end{subfigure}
    \begin{subfigure}[b]{0.48\columnwidth}
        \includegraphics[
            % inkscapelatex=false,
            width=\columnwidth,
        ]{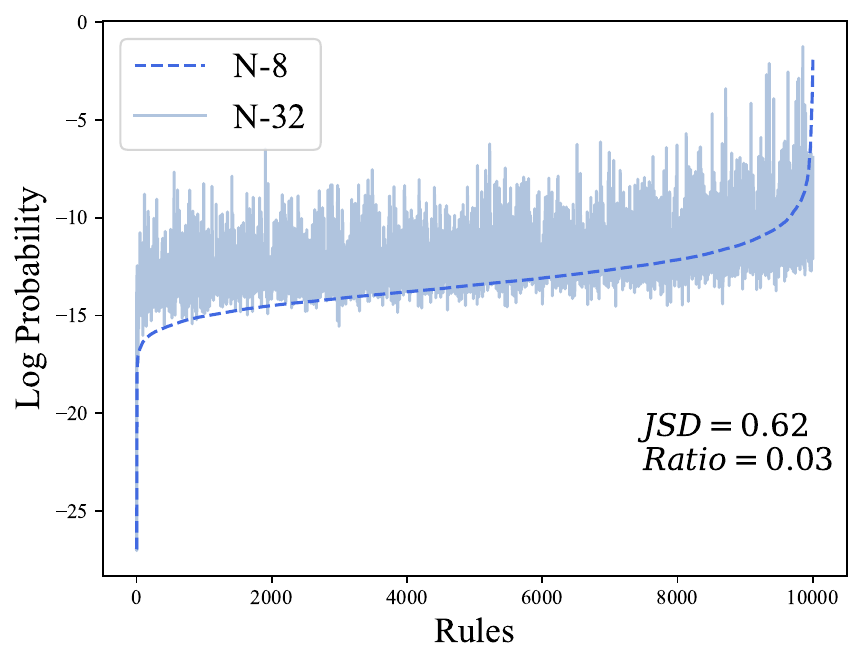}
        \caption{JSD = 0.62}
        \label{fig:jsd_d}
    \end{subfigure}
    \begin{subfigure}[b]{0.48\columnwidth}
        \includegraphics[
            % inkscapelatex=false,
            width=\columnwidth,
        ]{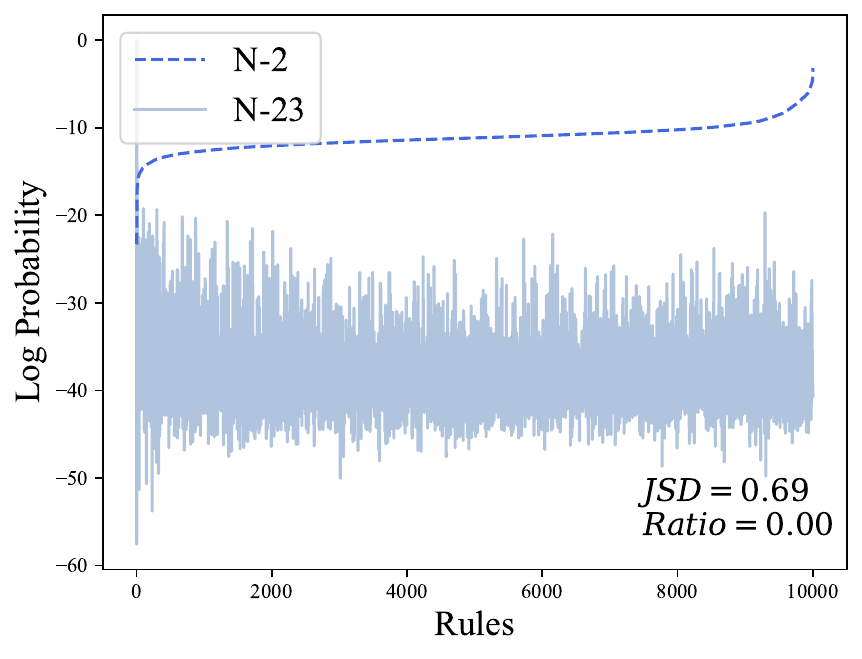}
        \caption{JSD = 0.69 (dissimilar)}
        \label{fig:jsd_e}
    \end{subfigure}
    \caption{Histogram that divided based on \textit{Jensen-Shannon Divergence} in (a). The histogram shows how many pairs of rule probability distributions fall into each bin. Blue and orange represent N-PCFG and ours. (b)-(e) provide examples from each bin of the histogram on N-PCFGs, showing the overlap of their utilized rules. Each blue line represents a different symbol.}
    \label{fig:example_collapse}
\end{figure}

Figure~\ref{fig:example_collapse} illustrates evidence for PDC in N-PCFGs.
In Figure~\ref{fig:example_collapse}(a), N-PCFGs reveal that while some rule distributions are distinct from each other, a large portion remains highly similar.
In contrast, our proposed approach yields a distribution more concentrated around 0.7, indicating that most rule distributions are more distinct.
Figure~\ref{fig:example_collapse}(b)–(d) show the sorted log probabilities for rules of two nonterminal symbols for each scale of JSD in N-PCFGs.
To quantify the diversity of rule usage, we compute the ratio of overlapping rules within the top 90\% of cumulative probability mass to the total number of rules contributing to that mass.
The figure for low JSD (e.g., 0.2) shows a high overlap ratio of 0.71, indicating limited distributional variation.
As JSD increases, this overlap ratio steadily declines, approaching zero near a JSD of 0.7, where rule probabilities differ substantially.
Therefore, N-PCFGs with fewer distinct distributions suffer from PDC, which limits their ability to fully utilize model capacity, suggesting potential for further utilization gain.

\section{Collapse-Relaxing Neural Parameterization}

In this section, we investigate the causes of PDC throughout all stages of neural parameterization. We discuss 1) symbol embeddings and their mapping to latent representations with limited expressiveness as a neural network in Section~\ref{sec:sym_emb}, 2) mapping latent representations to rule probability distributions entangled by embedding scale factors across symbols in Section~\ref{sec:latent_repr}, and 3) training dynamics causing representation collapse in Section~\ref{sec:training_dynamics}.
Then, we discuss the strategies to solve each issue, and propose an implementation on N-PCFG. 

\subsection{Symbol Embedding to Latent Representation}\label{sec:sym_emb}
%\subsection{Low Capacity of Parameterizer}
%\subsection{Small Parameterizer에 의한 최대 표현력의 한계}
%잠재 표현력은 Collapse를 어떻게 유발하는가?
\paragraph{Less Expressive Embedding and Layers}
Existing models typically employ small embedding sizes and shallow networks for neural parameterization. In particular, the parameterization of binary rules often relies on a single linear layer, which is a linear transformation of parent symbol embeddings to logits for rule probabilities, notated as: 
$T:\mathbf{R}^{emb}\rightarrow\mathbf{R}^{rules}$. 
When $emb < rules$ where $rules = \| N\cup P\|^2$, the mapping is surjective onto a subspace of the rule probability distribution space. As a result, certain rule probability distributions become unrepresentable given the limited dimensionality of the parent embeddings. When an optimal grammar requires those unrepresentable distributions, the trained grammar confined to the subspace becomes a projection of the optimum, exposed to losing distinction of the projected distributions, which is PDC effect. 
\paragraph{Solution}
The solution to this problem is relatively straightforward: increasing the dimensionality of symbol embeddings and the number of network layers. While this can be achieved through simple hyperparameter tuning, the following two issues may hinder performance gains, which requires a method that addresses them simultaneously.

%WORKING POINT
\subsection{Latent Representation to Probability Distribution}\label{sec:latent_repr}
%\subsection{Penultimate Representation Interference}
% Penultimate layer에서 발생하는 문제제
\paragraph{Entangled Children Scale Across Rules}
% \paragraph{Problem}
% 문제가 무엇인가?
The next step of neural parameterization also causes PDC due to the shared latent representations of children across different rules.
This step involves computing the inner product between the parent and child latent representations, which is then passed to the penultimate layer to generate logits for the softmax function representing the rule probability distribution.
The inner product is given by $\mathbf{u}_{BC}^{\mathsf{T}}\mathbf{w}_{A}$, where $\mathbf{w}_{A}$ and $\mathbf{u}_{BC}$ denote the latent representations of the parent and children, respectively, as shown in equation~(\ref{eq:binary_param}). The resulting rule probability can be rewritten in terms of the inner product as:
\begin{equation}\label{eq:inner_prod}
    \pi_{A\rightarrow BC} = \text{softmax}(\|\bf{w}_{A}\|\|\bf{u}_{BC}\|\cos\theta_{\bf{w}_{A},\bf{u}_{BC}})
\end{equation}
A notable point is that the scale of the children representation is shared across many parent rules.
For instance, the penultimate representations of rules $A \rightarrow BC$ and $A' \rightarrow BC$, involving different parent symbols, both include the shared children scale $\|\mathbf{u}_{BC}\|$.
This shared scale entangles the distributions of different rules, so a substantial increase in the children scale during training for one rule induces a corresponding increase for other rules involving the same children.
More critically, this scale entanglement manifests across all parent symbols that share the same children, occurring consistently for each children in the grammar.
As a result, a greater reliance on the children scale leads to convergence toward more similar rule distributions, thus causing PDC.

\paragraph{Solution}
% interference를 피하기 위해서는 shared factor를 어떻게 제어하는가?
To mitigate PDC caused by entangled children scales, a simple yet effective solution is to normalize the scale. By fixing the magnitude as a constant, the rule probability distribution becomes invariant to scale, preventing unintended influence from unrelated rules while preserving the learnability of child representations. 
In contrast, normalizing parent representations is generally discouraged, as each parent defines a distinct probability distribution. Normalization may unnecessarily reduce variance, limiting expressiveness.

% As previous discussed, the primary factor in neural PCFGs is the scale of children embedding.
% Therefore, we use normalized children embedding to exclude the scale information for inner product with parent representation in penultimate layer.
% Normalization for children embedding remove dependence for scale of children embedding, which avoid probability distribution collapse by relaxing the convergence bias.
% One thing to note is that we have to not normalize parent representations. This is because the scale of parent expression acts as a role in the variance of the probability distribution. Therefore the scale of parent representation $\|\bf{w}_{A}\|$ controls the smoothness of probability distribution. If we normalize the scale of parent representation, we always smoothed probability distribution.

\subsection{Training Dynamics}\label{sec:training_dynamics}

\paragraph{Bias to Specific Distributions}
During training with neural parameterization, two phenomena frequently occur: 1) gradient explosion that training is disrupted by excessively large gradients and 2) dying ReLU that activations passing through the ReLU function are mapped to zero.
These effects push the probability distribution toward specific forms, causing PDC that limit the expressiveness for more diverse distributions.
First, gradient explosion (GE) rapidly increases the scale of parent representations, resulting in an extremely sharp, one-hot-like distribution. This shift from a more diverse distribution to a near-deterministic one is a form of PDC. GE also easily increases the scale of children representations, which amplifies the previously discussed PDC, further concentrating probability mass on a few specific rules. 
To mitigate this, N-PCFGs often apply residual connections and gradient clipping. However, residual blocks still suffer from GE, and gradient clipping only partially alleviates the problem while potentially hindering the learning of other weights~\cite{Zhang2019FixupIR}. Thus, these methods offer limited effectiveness in resolving the core issue.
Second, dying ReLU causes most dimensions of the parent representation to become zero. In this case, as shown in equation~(\ref{eq:inner_prod}), the norm $|\mathbf{w}_{A}|$ approaches zero, causing the probabilities for all children representations to converge to equal values. Consequently, the grammar converges to a uniform probability distribution.

\paragraph{Solution}
% 불안정성 해결을 위해 어떤 방법을 선택했는가?
There are various strategies to mitigate gradient exploding and the dying ReLU problem. In this work, we adopt two simple yet effective methods to demonstrate that resolving these issues alleviates PDC:
1) We replace the ReLU activation with GELU~\cite{Hendrycks2016GaussianEL}, which offers two advantages. First, GELU allows activation in the negative domain, helping prevent the dying ReLU issue. Second, in the mean-field perspective, GELU exhibits a wider edge-of-chaos regime than ReLU, which helps stabilize gradients and mitigate explosion.
2) We remove residual connections for two reasons. First, residual networks without normalization still suffer from gradient explosion~\cite{Zhang2019FixupIR}. Second, even with normalization, residual blocks often suffer from scale imbalance between the main and residual paths~\cite{Hayou2020StableR}.

% There are various solutions to solve gradient exploding and dying ReLU issue. However we apply two simple methods to show that solving two issues solves collapse:
% 1) We use GELU~\cite{Hendrycks2016GaussianEL} as activation function instead of ReLU function, which have two advantages. First, GELU function allow to utilize negative interval, which make to avoid Dying ReLU issue. Second, GELU form wider edge-of-chaos area than ReLU in perspective of mean-field, which also release exploding gradient.
% 2) We do not use residual connection, which have two reasons. First, residual network without normalization still suffer from exploding gradient~\cite{Zhang2019FixupIR}.
% Even if we apply normalization techniques for residual blocks, imbalance of scale between main path and residual path is occured~\cite{Hayou2020StableR}.
% RMS Normalization을 왜 사용하는지가 불투명함.
% RMS Normalization~\cite{Zhang2019RootMS}을 사용한다. 이는 RMS normalization이 전체 embedding dimension에 걸친 고른 정보를 확인하는데에 더 유리하기 때문이다.

\subsection{Implementation of Rule Parameterization}
% 전략들을 실제 모델에 어떻게 적용했는가?
To use the proposed strategies, we present our redesign of the neural parameterization in N-PCFGs. 
% In this section, we present our redesign of the neural parameterization in Neural PCFGs based on the all proposed strategies. 
As described, the neural parameterization consists of three independent neural networks, each modeling one of subsets of the rule probability distributions: root rules ($S \rightarrow A$), binary rules ($A \rightarrow BC$), and unary rules ($T \rightarrow \omega$). We apply distinguished strategies to each network according to the characteristics of the corresponding distribution set.
% In this chapter, we introduce how we re-design the neural parameterization of Neural PCFGs based on proposed strategies. As we explained in Chapter 2.2, neural parameterization represent each set of rule probability distributions, $ROOT\rightarrow A$, $A\rightarrow BC$, $T\rightarrow\omega$ by using independent three neural networks.
% We apply different strategies according to the properties of each set of probability distributions.

\paragraph{Root Rules} 
The root rule probability distribution consists of a single distribution with a pseudo-root symbol as the parent.
In other words, it does not conflict with any other probability distribution.
Therefore, it does not require high expressive power, and we use a single fully connected layer to reduce computational cost.
% Root rule probability distribution is composed by only one probability distribution that have pseudo root symbol as parent.
% In other words, root rule probability distribution do not have other conflicting probability distribution.
% Therefore, root rule distribution is not need to many expressive power to represent. So, we use single fully-connected layer to reduce computational cost.
\begin{equation}\label{eq:new_root_param}
    \pi_{S\rightarrow A}=\frac{\exp(\mathbf{u}^{\mathsf{T}}_{A}\mathbf{w}_{S})}{\sum_{A'\in\mathcal{N}}\exp(\mathbf{u}^{\mathsf{T}}_{A'}\mathbf{w}_{S})}
\end{equation}

\paragraph{Binary and Unary Rules} 
Parameterizing binary and unary rule probability distributions is more complex than the root rule ($S \rightarrow NT$).
Binary rules require $|N|$ distinct probability distributions over $|N \cup U|^2$ possible child combinations, and unary rules require $|P|$ distinct distributions over $|\Sigma|$ terminals.
To learn distinct distributions for each parent symbol, the neural parameterization must have sufficient expressive power.
Accordingly, we compute the binary and unary rule distributions using the following equations:
% Parameterizing binary and unary probability distributions are more complex than $S\rightarrow NT$. They have $\| N\|$ probability distributions for each $\| N\cup U\|^2$ children rules, and $\|P\|$ probability distribution for each $\|\Sigma\|$ children. 
% To train different probability distribution for each parent symbols, we need enough expressive power for neural parameterization.
% Therefore, we calculate the binary and unary rules by the following equations:
\begin{align}
    \pi_{A\rightarrow BC}&=\frac{\exp(\|\mathbf{v}_A\|\cos\theta_{\mathbf{u}_{BC},v_{A}})}{\sum_{B'C'\in\mathcal{M}}\exp(\|\mathbf{v}_A\|\cos\theta_{\mathbf{u}_{B'C'},v_{A}})},\label{eq:new_binary_param}\\
    \pi_{T\rightarrow \omega}&=\frac{\exp(\|\mathbf{v}_{T}\|\cos\theta_{\mathbf{u}_{\omega},\mathbf{v}_{T}})}{\sum_{\omega'\in\Sigma}\exp(\|\mathbf{v}_{T}\|\cos\theta_{\mathbf{u}_{\omega'},\mathbf{v}_{T}})}\label{eq:new_unary_param}
\end{align}
where $\mathbf{v}_{A}$, $\mathbf{v}_{T}$ indicates the representations for parent symbols calculated with the equation:
\begin{equation}
    \mathbf{v}_{A}=\text{GELU}(\text{RMSNorm}(\mathbf{w}^{\mathsf{T}}\mathbf{w}_{A}))
\end{equation}

\section{Experiments}

\subsection{Settings}
\paragraph{Datasets}
% Dataset은 뭘 썼는가?
We evaluate the performance of models for constituency parsing task using the Penn TreeBank (PTB)~\cite{marcus1994penn} dataset for English, the Penn Chinese TreeBank (CTB)~\cite{xue2005penn} and the SPMRL dataset~\cite{seddah-etal-2014-introducing} for the other eight languages.
% Dataset에 대한 preprocessing은 어떻게 진행했는가?
We use the same preprocessing as \citet{yang2022dynamic} and \citet{park-kim-2024-structural}. We use the same train / development / test split as \citet{yang2022dynamic} for PTB, CTB and SPMRL dataset.
\paragraph{Hyperparameters}
% Hyperparameters는 어떻게 설정되었는가?
We train for 30 epochs with a batch size of 4 and use the same hyperparameters as N-PCFGs~\cite{kim2019compound}, except that we do not apply curriculum learning and gradient clipping.
For curriculum learning, TN-PCFGs~\cite{yang2021pcfgs} reported no performance benefit, and our own preliminary tests similarly showed no impact. It is also not used in TN-PCFGs, Rank PCFGs, or SimplePCFGs.
As for gradient clipping, we do not apply it because it hinders the learning of weights, as previously discussed in Section~\ref{sec:training_dynamics}.
Additionally, we use RMS Normalization~\cite{Zhang2019RootMS} after the activation function in CRNP to stabilize the forward path.
We run only four runs for each model.
% TODO: 이 부분 수정 필요
We follow the ratio of 1:2 between nonterminals and preterminals. The details for the dataset and training are in Appendix~\ref{appendix:exp_details}
We primarily evaluate parsing performance of grammars with MBR decoding~\cite{yang2021pcfgs}.

\begin{table}[!ht]
    \small
    \centering
    \begin{tabular}{clccc}
        \toprule
        \multirow{2}{*}[-0.5ex]{\textbf{Type}} & \multirow{2}{*}[-0.5ex]{\textbf{Model}} & \multirow{2}{*}[-0.5ex]{$|N|$} & \multicolumn{2}{c}{\textbf{S-F1}} \\
        \cmidrule{4-5}
        &&& \textbf{Mean} & \textbf{Max} \\
        \midrule
        &N-PCFG* & 30 & $\meanstd{52.3}{2.3}$ & 55.8 \\
        &C-PCFG* & 30 & $\meanstd{56.3}{2.1}$ & 60.1 \\
        &TN-PCFG* & 30 & $\meanstd{51.4}{4.0}$ & 55.6 \\
        &TN-PCFG* & 250 &  $\meanstd{57.7}{4.2}$ & 61.4\\
        Original & Rank PCFG$^{\dagger}$ & 30 & $\meanstd{51.2}{3.1}$ & - \\
        &Rank PCFG$^{\dagger}$ & 4500 & 64.1 & - \\
        &SN-PCFG$^{\ddagger}$ & 128 & $\meanstd{51.1}{4.1}$ & - \\
        &SN-PCFG$^{\ddagger}$ & 4096 & $\meanstd{65.1}{2.1}$ & - \\
        & CRNP (Ours) & 30 & $\meanstd{46.2}{5.2}$ & 52.7 \\
        \midrule
        Upper- & N-PCFG+PF$_g$  & 30 & $\meanstd{72.6}{0.7}$ & 73.4 \\
        bound & CRNP+PF$_g$ & 30 & $\meanstd{\textbf{73.7}}{0.3}$ & 74.2 \\
        \midrule
        PF & N-PCFG+PF & 30 & $\meanstd{68.3}{0.2}$ & 68.7 \\
        (fixed- & Rank PCFG+PF & 30 & $\meanstd{67.4}{0.9}$ & 68.4 \\
        NT size) & CRNP+PF & 30 & $\meanstd{\textbf{69.4}}{0.3}$ & 69.7 \\
        \midrule
        PF (no- & Rank PCFG+PF & 4500 & $\meanstd{69.6}{0.6}$ & 70.3 \\
        limit) & CRNP+PF & \textbf{90} & $\meanstd{\textbf{70.2}}{0.5}$ & 70.9 \\
        % \hline
        % N-PCFG+PF (Gold) & 30 & $\meanstd{72.6}{0.7}$ & 73.4 \\
        % CRNP+PF (Gold) & 30 & $\meanstd{73.7}{0.3}$ & 74.2 \\
        % \hline
        % N-PCFG+PF (Pretrained) & 30 & $\meanstd{68.3}{0.2}$ & 68.7 \\
        % Rank PCFG+PF (Pretrained) & 30 & $\meanstd{67.4}{0.9}$ & 68.4 \\
        % Rank PCFG+PF (Pretrained)& 4500 & $\meanstd{69.6}{0.6}$ & 70.3 \\
        % CRNP+PF (Pretrained) & 30 & $\meanstd{69.4}{0.3}$ & 69.7 \\
        % CRNP+PF (Pretrained) & 90 & $\meanstd{70.2}{0.5}$ & 70.9 \\
        \bottomrule
    \end{tabular}
    \caption{Performance measured by unlabeled sentence-F1 (S-F1) in English constituent parsing on PTB. $|N|$ represents the number of nonterminals. *, $\dagger$, and $\ddagger$ indicate reported results from \citet{yang2021pcfgs}, \citet{yang2022dynamic}, and \citet{liu_simple_2023}, respectively.
    Parse-focusing PF$_g$ uses gold parse trees, while PF uses parse trees induced from pretrained models in an unsupervised setting without extra data.}
    \label{tab:final-performance}
\end{table}

\paragraph{Parse-Focusing}

\begin{table*}[!ht]
    \centering
    \begin{adjustbox}{max width=\textwidth}
    \begin{tabular}{lcccccccccccc}
    \toprule
        \textbf{Model} & $|N|$ & \textbf{Basque} & \textbf{Chinese} & \textbf{English} & \textbf{French} & \textbf{German} & \textbf{Hebrew} & \textbf{Hungarian} & \textbf{Korean} & \textbf{Polish} & \textbf{Swedish} & \textbf{Mean rank}\\
    \midrule
        N-PCFG* & 30 &
        $\meanstd{35.1}{2.0}$ & $\meanstd{26.3}{2.5}$ & $\meanstd{52.3}{2.3}$ & $\meanstd{45.0}{2.0}$ & $\meanstd{42.3}{1.6}$ & $\meanstd{45.7}{2.2}$ & $\meanstd{43.5}{1.2}$ & $\meanstd{28.4}{6.5}$ & $\meanstd{43.2}{0.8}$ & $\meanstd{17.0}{9.9}$ & 6.2 \\
        C-PCFG* & 30 &
        $\meanstd{36.0}{1.2}$ & $\meanstd{38.7}{6.6}$ & $\meanstd{56.3}{2.1}$ & $\meanstd{45.0}{1.1}$ & $\meanstd{43.5}{1.2}$ & $\meanstd{45.2}{0.5}$ & $\mathbf{\meanstd{\color{blue}44.9}{1.5}}$ & $\meanstd{30.5}{4.2}$ & $\meanstd{43.8}{1.3}$ & $\meanstd{33.0}{15.4}$ & 5.0 \\
        CRNP+PF (Ours) & 30 & \underline{$\meanstd{{\color{blue}46.1}}{0.2}$} & $\meanstd{\color{blue}45.8}{1.7}$ & $\meanstd{\color{blue}69.4}{0.3}$ & $\meanstd{\color{blue}50.4}{0.1}$ & $\meanstd{\color{blue}47.9}{0.2}$ & $\bf{\meanstd{\color{blue}49.5}{0.3}}$ & $\meanstd{43.9}{0.1}$ & $\meanstd{\color{blue}41.9}{0.8}$ & $\meanstd{\color{blue}45.8}{0.9}$ & $\meanstd{\color{blue}34.0}{0.9}$ & \color{blue}2.9 \\
    \midrule        
        TN-PCFG* & 250 &
        $\meanstd{36.0}{3.0}$ & $\meanstd{39.2}{5.0}$ & $\meanstd{57.7}{4.2}$ & $\meanstd{39.1}{4.1}$ & $\meanstd{47.1}{1.7}$ & $\meanstd{39.2}{10.7}$ & $\meanstd{43.1}{1.1}$ & $\meanstd{35.4}{2.8}$ & $\mathbf{\meanstd{\color{blue}48.6}{3.1}}$ & $\mathbf{\meanstd{\color{blue}40.0}{4.8}}$ & 4.5 \\
        Rank PCFG$^{\dag}$ & 4500 & 
        $\meanstd{38.4}{7.3}$ & $\meanstd{31.0}{8.4}$ & $\meanstd{59.6}{7.7}$ & $\meanstd{43.9}{3.1}$ & $\meanstd{48.0}{1.4}$ & $\meanstd{46.2}{4.1}$ & $\meanstd{42.2}{0.7}$ & $\meanstd{31.5}{4.0}$ & $\meanstd{41.6}{4.3}$ & $\mathbf{\meanstd{\color{blue}40.0}{0.6}}$ & 4.7 \\
        Rank PCFG+PF$^{\dag}$ & 4500 &
        $\meanstd{45.9}{0.3}$ & \underline{$\meanstd{46.1}{0.9}$} & \underline{$\meanstd{69.7}{0.9}$} & \underline{$\meanstd{50.5}{0.5}$} & $\mathbf{\meanstd{\color{blue}49.1}{0.3}}$ & $\mathbf{\meanstd{\color{blue}49.5}{0.2}}$ & $\meanstd{43.7}{0.2}$ & \underline{$\meanstd{42.1}{0.3}$} & \underline{$\meanstd{47.9}{0.3}$} & $\meanstd{33.4}{1.2}$ & \underline{2.4} \\
        CRNP+PF (Ours) & \color{blue}90 & $\mathbf{\meanstd{\color{blue}46.3}{0.6}}$ & $\mathbf{\meanstd{\color{blue}46.6}{1.0}}$ & $\mathbf{\meanstd{\color{blue}70.2}{0.5}}$ & $\mathbf{\meanstd{\color{blue}51.4}{0.3}}$ & \underline{$\meanstd{48.7}{0.2}$} & $\bf{\meanstd{\color{blue}49.5}{0.5}}$ & \underline{$\meanstd{\color{blue}44.1}{0.2}$} & $\mathbf{\meanstd{\color{blue}42.3}{0.3}}$ & $\meanstd{45.7}{0.4}$ & \underline{$\meanstd{34.4}{0.6}$} & \textbf{\color{blue}1.7} \\
    % \hline
    %     Oracle & 69.82 & 85.95 & 84.25 & 76.47 & 64.91 & 85.79 & 63.79 & 99.14 & 74.75 & 72.09 \\
    \bottomrule
    \end{tabular}
    \end{adjustbox}
    \caption{Performance (S-F1 score and mean rank) in multilingual parsing on PTB, CTB, and SPMRL datasets. (*,$\dagger$ :reported results in~\citet{yang2021pcfgs},~\citet{park-kim-2024-structural}, respectively, $|N|$: number of nonterminals). The bold represent the best performance and the underline represent the second performance. Mean rank represents the average rank of each model across all languages. Blue text indicates the best result in each division, either with NT fixed to 30 or with NT unrestricted.}
    \label{tab:multilingual-evaluation}
\end{table*}

% % Parse-Focusing의 적용과 적용하려는 이유
% By mitigating not only probability distribution collapse but also previously known potential issues, namely structural optimization ambiguity (SOA) and structural simplicity bias (SSB), we reveal the upper-bound performance achievable by neural grammars in unsupervised neural grammar induction. To this end, we apply Parse-Focusing to address SOA and SSB~\cite{park-kim-2024-structural}.
% % Parse-Focusing으로 보고자 하는 결과
% % focusing-bias로부터 올 수 있는 성능을 배제하고, 이론적 성능의 upper bound를 확인할 수 있다.
% For parse-focusing, we use two different focusing biases, binarized gold parse trees and induced parse trees from pretrained models (TN-PCFG~\cite{yang2021pcfgs}, NBL-PCFG~\cite{yang2021neural} and Structformer~\cite{shen2021structformer}) provided by ~\citet{park-kim-2024-structural}. The parse-focusing with gold parse trees makes the model is trained in the environment that similar with supervised learning.

N-PCFGs already suffer from known potential issues, namely Structural Optimization Ambiguity (SOA) and Structural Simplicity Bias (SSB)~\cite{park-kim-2024-structural}, which may overshadow the effects of mitigating PDC.
To clearly analyze these effects, we apply Parse-Focusing (PF)~\cite{park-kim-2024-structural} to N-PCFGs and CRNP.
The Parse-Focusing method was originally proposed to utilize the parse trees induced by pretrained unsupervised models (TN-PCFG~\cite{yang2021pcfgs}, NBL-PCFG~\cite{yang2021neural}, and Structformer~\cite{shen2021structformer}) in the unsupervised setting, and we follow this setup.
Additionally, we use the gold parse trees to reveal the experimental upper bound achievable in UNGI.
In this setting, we assume that the gold parse trees provide the ideal focusing bias and thus yield the best performance. This corresponds to the supervised setting. 

\subsection{Performance Results}

\paragraph{Performance in English}
Table~\ref{tab:final-performance} presents the unlabeled sentence-level F1 (S-F1) scores on the PTB dataset for various baselines and our model with the proposed parameterization (CRNP).
CRNP without PF shows a lower mean and higher variance in S-F1 scores compared to conventional neural parameterization, due to the presence of SOA and SSB.
When PF is applied using parse trees induced from pretrained models on the same dataset, CRNP+PF consistently outperforms N-PCFG+PF.
CRNP+PF with 90 nonterminals demonstrates significant performance improvements across most baselines, achieving slightly higher performance than Rank PCFG+PF with 4500 nonterminals, a grammar that is 50 times larger.
Notably, when both models use 30 nonterminals, CRNP+PF shows a substantial performance gain over Rank PCFG+PF.
Furthermore, when gold parse trees are used for PF, CRNP+PF with just 30 nonterminals surpasses N-PCFG+PF, demonstrating a stronger upper bound.

\paragraph{Performance in Multilingual}

% Table 2는 무엇인가?
Table~\ref{tab:multilingual-evaluation} presents the S-F1 scores for parsing performance on PTB, CTB, and SPMRL datasets.
Overall, the proposed method, CRNP+PF, achieves substantial improvements across most languages, including Basque, Chinese, English, French, Hebrew, and Korean, and ranks second for German, Hungarian, and Swedish, resulting in the best average rank (1.7) among all models.
These results highlight the effectiveness of CRNP across diverse language datasets.
Under a fixed grammar size of 30 nonterminals, CRNP+PF consistently outperforms other models.
Even under the unrestricted setting, it achieves the highest scores in most cases, while still using a highly compact grammar with only 90 nonterminals.

\begin{table}[htbp]
    \centering
    \begin{adjustbox}{max width=0.8\columnwidth}
    \begin{tabular}{lccc}
    \toprule
        \multirow{2}{*}[-0.5ex]{$|N|$} & \multicolumn{3}{c}{S-F1} \\
        \cmidrule{2-4}
        & Rank PCFG & Rank PCFG+PF & CRNP+PF \\ 
    \midrule
        1 & $\meanstd{39.7}{0.0}$ & $\meanstd{44.3}{0.1}$ & $\meanstd{45.0}{0.1}$ \\
        5 & $\meanstd{41.6}{9.6}$ & $\meanstd{60.4}{1.9}$ & $\meanstd{59.4}{3.0}$ \\
        15 & $\meanstd{43.2}{0.8}$ & $\meanstd{65.5}{0.7}$ & $\meanstd{67.5}{0.7}$ \\
        30 & $\meanstd{51.2}{3.1}$ & $\meanstd{67.4}{0.9}$ & $\meanstd{69.4}{0.3}$ \\
        60 & - & - & $\meanstd{70.0}{0.2}$ \\
        90 & - & - & $\meanstd{70.2}{0.5}$ \\
        250 & $\meanstd{54.3}{3.9}$ & $\meanstd{69.7}{0.9}$ & - \\
        4500 & $\meanstd{57.4}{6.0}$ & $\meanstd{69.6}{0.7}$ & - \\
    \bottomrule
    \end{tabular}
    \end{adjustbox}
    \caption{Performance (S-F1) by grammar size to show grammar compactness.}
    \label{tab:num-of-sym}
\end{table}
\paragraph{Grammar Compactness}

% Table 3는 무엇인가?
Table~\ref{tab:num-of-sym} compares performance across grammar sizes for N-PCFG, Rank PCFG, and CRNP.
The results show that CRNP exhibits a steeper improvement and more accurate performance in S-F1 scores as grammar size increases, compared to the baseline models.
This demonstrates that CRNP effectively leverages larger grammars by mitigating the bottleneck limitations of neural parameterization.

% Table~\ref{tab:num-of-sym} compare the performance according to the grammar size on N-PCFG, Rank PCFG and CRNP.
% % 더 작은 grammar로도 많은 symbols를 사용하는 모델만큼의 성능 확보 가능
% The results show the steep increase of the S-F1 scores according to the grammar size on the CRNP than existing models.
% % 무엇을 말하고자 하는가?
% We show that CRNP achieve the increase of performance according to the grammar size by relaxing the limitation from the bottleneck.

\begin{table}[!h]
    \centering
    \small
    \begin{tabular}{lc}
    \hline
    Model & S-F1 \\
    \hline
    % CRNP$_{-\text{GELU}-\text{RMSNorm}-\text{NormalizedEmb}}$ & \\
    
    CRNP+PF$_{-\text{GELU}-\text{NormalizedEmb}}$ & $\meanstd{72.4}{0.4}$ \\
    CRNP+PF$_{-\text{GELU}+\text{ReLU}}$ & $\meanstd{72.9}{0.2}$ \\
    % CRNP$_{-\text{RMSNorm}}$ & $\meanstd{73.6}{0.6}$ \\
    CRNP+PF$_{-\text{NormalizedEmb}}$ & $\meanstd{73.5}{0.2}$ \\
    CRNP+PF & $\meanstd{\textbf{73.7}}{0.3}$ \\
    \hline
    \end{tabular}
    \caption{Ablation studies on upper bound performance ($_{-\text{GELU}+\text{ReLU}}$: using ReLU instead of GELU, 
    % $_{\text{-RMSNorm}}$: no normalization, 
    $_{\text{-NormaliedEmb}}$: no normalization)
    % represent the model that do not use normalized children embedding in penultimate calculation.
    }
    \label{tab:ablation}
\end{table}
% \subsection{In-Depth Analysis}
\paragraph{Ablation Studies}
For the ablation study, we examine the individual contributions of GELU activation and children embedding scale normalization, which are two core components of CRNP, to the upper bound without interference from unsupervised learning environments.
Table~\ref{tab:ablation} presents S-F1 scores under different configurations.
The removal of either component consistently degrades performance, indicating that both are essential to the effectiveness of the proposed method.
The ablation study on varying embedding dimensions, another core component, requires a more in-depth analysis. Therefore, we discuss it separately in Section~\ref{sec:in_depth}.

\subsection{In-Depth Analysis}\label{sec:in_depth}
\paragraph{Mitigation of Probability Distribution Collapse}

\begin{table}[!h]
    \centering
    \begin{adjustbox}{max width=0.8\columnwidth}
    \begin{tabular}{clccc}
    \hline
    \multirow{2}{*}{Rule Type} &\multirow{2}{*}{Model} &\multicolumn{3}{c}{GPJ with varying $|N|$} \\
    \cline{3-5}
     && 30 & 60 & 90 \\
    \hline
    % \multicolumn{4}{c}{Binary rules}\\
    % \hline
    \multirow{3}{*}{Binary}&N-PCFG & 0.661 & 0.661 & 0.657 \\
    &N-PCFG+PF & 0.647 & 0.645 & 0.634 \\
    &CRNP+PF & 0.647 & 0.657 & 0.662 \\ 
    \hline
    % \multicolumn{4}{c}{Unary rules}\\
    % \hline
    \multirow{5}{*}{Unary} &N-PCFG & 0.561 & 0.591 & 0.590 \\
    &TN-PCFG & 0.481 & - & - \\
    &Rank PCFG & 0.587 & - & - \\
    &N-PCFG+PF & 0.615 & 0.634 & 0.636 \\
    &CRNP+PF & 0.633 & 0.646 & 0.649 \\
    \hline
    \end{tabular}
    \end{adjustbox}
    \caption{Probability distribution collapse (measured by geometric mean of pairwise JSD (GPJ)) by the grammar size (nonterminal and preterminal size).}
    \label{tab:jsd}
\end{table}

% PDC의 완화를 확인
Beyond performance and compactness improvement, we validate that PDC is effectively reduced by our approach.
% JSD를 왜 사용하는가?
To evaluate the impact, we utilize the Jensen-Shannon Divergence (JSD), a symmetric and finite metric that measures the similarity between probability distributions.
The range of JSD is $(0, \ln{2}) \approx (0.00, 0.69)$. 
% JSD로 무엇을 알 수 있는가?
A high JSD indicates a greater difference between two probability distributions, while a value of zero indicates identical distributions.
To verify the overall tendency of PDC in grammars, we employ the geometric mean of pairwise JSD (GPJ) for rule probability distributions.
The GPJ reflects the diversity of the distributions, thereby implying the degree of capacity utilization.
We address the detailed definition of GPJ in Appendix~\ref{sec:measures}.
% 자료와 해석
In Table~\ref{tab:jsd}, CRNP exhibits high GPJ, indicating that the probability distributions in the grammar are distinct from one another.
In contrast, N-PCFGs show low GPJ for unary probability distributions while maintaining a comparable GPJ for binary distributions.
Similarly, TN-PCFGs and Rank PCFGs exhibit low GPJ for unary rules, which means they also suffer from PDC.
These results demonstrate the effectiveness of our approach in mitigating PDC and improving capacity utilization.

\paragraph{Embedding Size}
\begin{table}[!h]
    \centering
    \begin{adjustbox}{max width=0.8\columnwidth}
    \begin{tabular}{lccc}
    \hline
    \multirow{2}{*}{\shortstack{Embedding\\Size}} & \multicolumn{3}{c}{S-F1} \\
    \cline{2-4}
    & N-PCFG & Rank PCFG+PF & CRNP+PF \\
    \hline
    64      & 54.3 & 65.9 & 68.6%2 
    \\
    128     & 54.8 & 67.2 & 68.8%3 
    \\
    256     & 52.8 & 68.4 & 69.4%0 
    \\
    512     & 51.8 & 66.5 & 69.2%6 
    \\
    1024    & 50.6 & 66.5 & 69.2%2 
    \\
    2048    & 45.4 & 67.3 & 69.3 \\
    \hline
    \end{tabular}
    \end{adjustbox}
    \caption{Impact (S-F1) of the size of input embeddings and hidden states.}
    \label{tab:f1-emb}
\end{table}

We investigate the impact of model expressiveness by evaluating performance across varying symbol embedding and hidden state sizes, as shown in Table~\ref{tab:f1-emb}.
For N-PCFGs, S-F1 scores increase with embedding size up to 128 but decline beyond that, likely due to training dynamics that exacerbate PDC in larger networks.
In contrast, CRNP performance continues to improve with larger embeddings until saturation, which occurs once the expressive power matches the complexity of the target grammar. Beyond this point, further improvement would require increasing grammar size. 
% as the neural parameterization bottleneck has been reduced.
% We first demonstrate the performances by controlling the size of symbol embedding and hidden states of each layers as table~\ref{tab:f1-emb}.
% % 결과 분석 - existing model
% In N-PCFGs, the S-F1 scores increase in proportion of the embedding size until 128, however, performances decrease when the embedding size is bigger than 128. It is because that the probability distribution collapse by training dynamics is easy to appear when large networks.
% % proposed model
% While, the performance of CRNP increase according to the embedding size. the performance of CRNP is not increased more above the specific embedding size that satisfy expressive power that required by grammar size because the bottleneck from the neural parameterization is resolved enough. We have to increase grammar size to get more performance increase.

\paragraph{The Number of Layers}

\begin{table}[h]
    \centering
    \begin{adjustbox}{max width=0.8\columnwidth}
    \begin{tabular}{ccccc}
    \hline
     Layers 
     & 1
     & 2
     & 3
     & 4
     \\
     \hline
    S-F1
    & $\meanstd{68.3}{0.6}$ 
    & $\meanstd{69.2}{0.4}$ 
    & $\meanstd{69.4}{0.3}$ 
    & $\meanstd{69.2}{0.3}$ \\
    % The number of layers & S-F1    \\
    % \hline
    % 1  & $\meanstd{68.3}{0.6}$ \\
    % 2  & $\meanstd{69.2}{0.4}$ \\
    % 3  & $\meanstd{69.4}{0.3}$ \\
    % 4  & $\meanstd{69.2}{0.3}$ \\
    \hline
    \end{tabular}
    \end{adjustbox}
    \caption{Impact of (S-F1) the number of layers by the depth of neural parameterization layers.}
    \label{tab:f1-depth}
\end{table}
Another important factor in controlling network capacity is the number of layers.
Table~\ref{tab:f1-depth} reports performance changes as network depth increases in the neural parameterization.
Unlike N-PCFGs, our approach enhances model capacity by increasing the number of layers, thereby improving nonlinearity.
Consistent with the previous results from varying embedding size, performance saturates once sufficient expressiveness is achieved at two layers, with no further gains observed from additional depth.
% Another key factor for controlling network capacity is the number of layers.
% Table~\ref{tab:f1-depth} presents the performance change by network depth in the neural parameterization.
% Unlike N-PCFGs, our approach enhances capacity by increasing the number of layers and therefure the nonlinearity. 
% Just as the embedding size control, after the sufficient expressiveness is achieved with the two layers, the performance is saturated showing no more improvement with more layers. 
% One of the other component related to control the capacity of network is the number of layers.
% % Table 설명 (Table 완성 후 보강)
% Table~\ref{tab:f1-depth} shows the change of performance according to the number of layers in neural network used in neural parameterization.
% We also try to increase the capacity by increasing the nonlinearity of network with the increase of the number of layers unlike N-PCFGs.

\begin{table}[!h]
    \centering
    \small
    \begin{tabular}{lcc}
    \hline
    \multirow{2}{*}{Model} & \multicolumn{2}{c}{cosine similarity}    \\
    \cline{2-3}
     & binary & unary    \\
    \hline
    N-PCFG & 0.85 & 0.71 \\
    N-PCFG+PF & 0.91 & 0.78 \\
    CRNP+PF & 0.75 & 0.30 \\
    \hline
    \end{tabular}
    \caption{Average of cosine similarity among children representations used as inputs of the penultimate layer}
    \label{tab:f1-scale}
\end{table}
\paragraph{Children Scale}
% Description
To examine the effect of normalizing the scale of children representations, we evaluate their cosine similarity, as shown in Table~\ref{tab:f1-scale}. Normalization in CRNP significantly reduces similarity for both binary and unary rules, indicating more diverse distributions. This outcome is expected, as removing scale from the learnable parameters places greater emphasis on cosine similarity to capture distributional differences. These results suggest normalizing the children scale enhances representational diversity and encourages more varied rule probability distributions.

\begin{figure}[h!]
    \centering
    \fontsize{20}{10}\selectfont
    \includegraphics[
        width=5cm
    ]{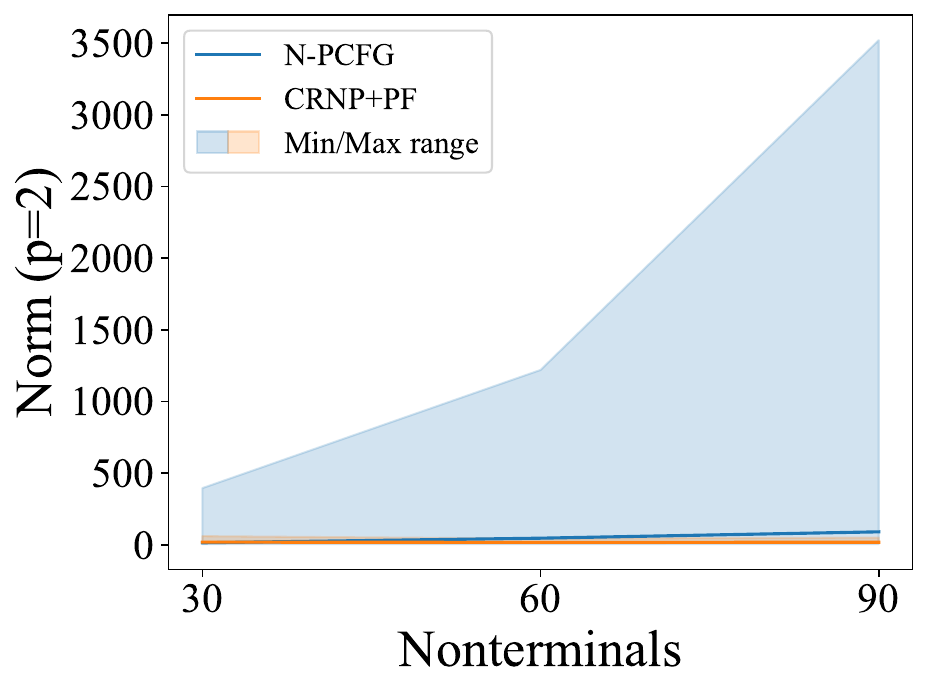}
    \caption{Gradient explosion evaluation reflected by embedding scales by nonterminals.}
    \label{fig:solving-PDC}
\end{figure}

\paragraph{Gradient Explosion}
We evaluate the scale of children representations to demonstrate the presence of gradient explosion and the effectiveness of our proposed method. Figure~\ref{fig:solving-PDC} shows the scale values before and after applying our approach. When the number of non-terminals is 30, both models exhibit comparable scales. However, as the number of non-terminals increases, the N-PCFG exhibits an exponential increase in scale values by the large maximum, indicating the onset of severe gradient explosion. In contrast, our method maintains consistently low scales and a narrow range, even with larger grammars. These results confirm that our approach effectively stabilizes the representation scale and mitigates critical PDC caused by sharing the scale factor in children representations.

% \begin{figure}[h!]
%     \centering
%     \includegraphics[
%         width=6cm
%     ]{rule_util.pdf}
%     \caption{The number of embeddings affected by dying ReLU.}
%     \label{fig:dying_ReLU}
% \end{figure}

\begin{table}[h!]
    \centering
    \begin{adjustbox}{max width=0.8\columnwidth}
    \begin{tabular}{ccccc}
    \hline
    Rule Type &Model & $|N|$ & Entropy & Ratio of zeros \\
    \hline
    % \multicolumn{4}{c}{Binary Rules} \\
    % \hline
    \multirow{6}{*}{Binary}&\multirow{3}{*}{N-PCFG} & 30 & 5.168 & 0.990 \\
    && 60 & 4.996 & 0.994 \\
    && 90 & 4.506 & 0.993 \\
    \cline{2-5}
    &\multirow{3}{*}{CRNP+PF} & 30 & 0.808 & 0.000 \\
    && 60 & 0.873 & 0.000 \\
    && 90 & 0.720 & 0.000 \\
    \hline
    % \multicolumn{4}{c}{Unary Rules} \\
    % \hline
    \multirow{6}{*}{Unary}&\multirow{3}{*}{N-PCFG} & 30 & 2.596 & 0.000 \\
    && 60 & 5.148 & 0.000 \\
    && 90 & 4.718 & 0.000 \\
    \cline{2-5}
    &\multirow{3}{*}{CRNP+PF} & 30 & 0.743 & 0.000 \\
    && 60 & 0.886 & 0.000 \\
    && 90 & 0.794 & 0.000 \\
    \hline
    \end{tabular}
    \end{adjustbox}
    \caption{Evaluation of the dying ReLU phenomenon. Entropy is computed from each nonterminal’s rule probability distribution, and the ratio from ReLU-generated representations. Both are averaged over all rules. Higher entropy indicates greater uniformity; the zero ratio reflects dying elements in the representations.}
    \label{fig:dying_relu}
\end{table}
    
\begin{figure*}[!h]
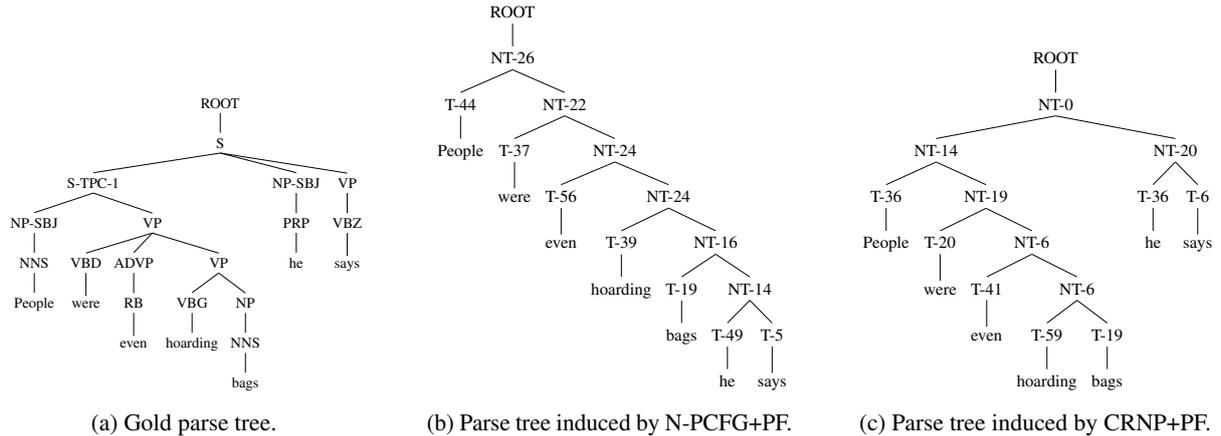

\centering
    \begin{subfigure}[b]{0.3\textwidth}
        \resizebox{\textwidth}{!}{
        \Tree [.ROOT
        [.S
          [.S-TPC-1
            [.NP-SBJ [.NNS People ] ]
            [.VP
              [.VBD were ]
              [.ADVP [.RB even ] ]
              [.VP
                [.VBG hoarding ]
                [.NP [.NNS bags ] ]
              ]
            ]
          ]
          [.NP-SBJ [.PRP he ] ]
          [.VP [.VBZ says ] ] ] ]
        }
        \caption{Gold parse tree.}
    \end{subfigure}
    \hfill
    \begin{subfigure}[b]{0.3\textwidth}
        \resizebox{\textwidth}{!}{
        \Tree [.ROOT
        [.NT-26
          [.T-44 People ]
          [.NT-22
            [.T-37 were ]
            [.NT-24
              [.T-56 even ]
              [.NT-24
                [.T-39 hoarding ]
                [.NT-16
                  [.T-19 bags ]
                  [.NT-14 [.T-49 he ] [.T-5 says ] ]
                ]
              ]
            ]
          ]
        ] ]
        }
        \caption{Parse tree induced by N-PCFG+PF.}
    \end{subfigure}
    \hfill
    \begin{subfigure}[b]{0.3\textwidth}
        \resizebox{\textwidth}{!}{
        \Tree [.ROOT
        [.NT-0
          [.NT-14
            [.T-36 People ]
            [.NT-19
              [.T-20 were ]
              [.NT-6 [.T-41 even ]
                [.NT-6 [.T-59 hoarding ] [.T-19 bags ] ]
              ]
            ]
          ]
          [.NT-20 [.T-36 he ] [.T-6 says ] ]
        ] ]
        }
        \caption{Parse tree induced by CRNP+PF.}
    \end{subfigure}
\caption{Comparison between the gold parse tree and the parse trees induced by the models (N-PCFG+PF and CRNP+PF) on the example sentence \textit{"People were even hoarding bags he says"}.}
\label{fig:qa_tree_comp}
\end{figure*}

\paragraph{Dying ReLU}
In Table~\ref{fig:dying_relu}, N-PCFG consistently exhibits higher entropy, indicating that its rule probability distributions are, on average, more uniform. However, high entropy does not necessarily imply the absence of inactive (dying) components in the representations. To assess this, we also measure the ratio of zero values. While N-PCFG shows a near-one ratio in binary rules, CRNP+PF maintains the zero rate. This suggests that N-PCFG suffers from the dying ReLU problem, whereas the proposed method effectively mitigates it. There is no dying ReLU phenomenon for binary rules, because N-PCFGs do not use an activation function to parameterize binary rules.

\paragraph{The Difference in Tree Structure}
For qualitative analysis, we provide example parse trees in Figure~\ref{fig:qa_tree_comp} to examine the influence of CRNP on parse structures.
N-PCFG+PF generates a right-binarized parse tree that differs from the gold by relying heavily on a few nonterminal symbols such as NT-22 and NT-24, whereas CRNP+PF reproduces the same structure as the gold by utilizing a wider variety of symbols.
From the perspective of symbol roles, CRNP+PF utilizes a broader range of symbols for S and VP than N-PCFG+PF, indicating a more diverse representation of intermediate sentence structures.
We further present additional examples in Appendix~\ref{appendix:trees}.

\section{Related Works}
% Overparameterization of Neural Grammar induction
\paragraph{Overparameterization of UNGI} 
% TN-PCFG / Rank PCFG / Simple PCFG
\citet{yang2021pcfgs} proposed the neural parameterization based on tensor decomposition, and demonstrated that overparameterization leads to better grammars with many symbols.
Moreover, \citet{yang2022dynamic} improved it based on factor graph grammars (FGGs).
Recently, \citet{liu_simple_2023} introduced the SimplePCFG formalism.
These show the performance improvements based on a large grammar size.
Conversely, we highlight the inefficiency in learning UNGIs caused by an implicit bottleneck, and demonstrate resolving this issue leads to compact and accurate grammars.

\paragraph{Collapse Problem}
In self-supervised learning (SSL), the collapse problem refers to representations with different semantics being mapped to very close positions.
\citet{9709917} verified complete collapse that representations converge into a trivial solution, and identified dimensional collapse that representations concentrate into sparse dimensions.
In another line of work, \citet{Papyan2020PrevalenceON} proposed neural collapse in which representations within the same class converge to their class means in classification tasks.
These studies focus on representations themselves. 
Whereas we focus on the collapse of probability distributions generated from representations and their negative effect, which directly limits capacity utilization of grammars.

\section{Conclusion}
In this paper, we address the issue of \textit{probability distribution collapse}, which creates a bottleneck in neural parameterization and limits the effective utilization of model capacity in unsupervised neural grammar induction. We analyze the underlying causes of this collapse, tracing training dynamics from symbol embeddings to the probability distributions, and propose a \textit{collapse-relaxing neural parameterization} to mitigate it. Our method improves the upper bound and overall accuracy of constituency parsing across English and multiple languages, while achieving significantly more compact grammars. This work underscores the often overlooked limitations of neural parameterization and reveals that large-capacity models are not strictly necessary for effective grammar learning. 

\section*{Acknowledgments}

This work was supported by the National Research Foundation of Korea (NRF) grant funded by the Korea government (MSIT) (No.2022R1A2C2012054, Development of AI for Canonicalized Expression of Trained Hypotheses by Resolving Ambiguity in Various Relation Levels of Representation Learning).

\section*{Limitations}
% 한계점 
Unsupervised neural grammar induction using N-PCFGs is still constrained by expensive computational cost. Therefore, extending our idea to a decomposed tensor-based method is important to scale up.
However, applying our method to models such as TN-PCFGs or Rank PCFGs is constrained by tensor decomposition.
While PDC may affect these models as well, further analysis is required to assess its impact and applicability.
In addition, we observed that CRNP+PF with 90 nonterminals performs slightly less effectively in a few languages (e.g., German and Polish).
However, language-specific differences in resolving PDC are out of the main scope of this work.
Therefore, such issues should be further investigated in future work.

% End

\bibliography{custom}

\begin{thebibliography}{29}
\expandafter\ifx\csname natexlab\endcsname\relax\def\natexlab#1{#1}\fi

\bibitem[{Buhai et~al.(2019)Buhai, Halpern, Kim, Risteski, and Sontag}]{Buhai2019EmpiricalSO}
Rares-Darius Buhai, Yoni Halpern, Yoon Kim, Andrej Risteski, and David~A. Sontag. 2019.
\newblock \href {https://api.semanticscholar.org/CorpusID:219687014} {Empirical study of the benefits of overparameterization in learning latent variable models}.
\newblock In \emph{International Conference on Machine Learning}.

\bibitem[{Drozdov et~al.(2020)Drozdov, Rongali, Chen, O{'}Gorman, Iyyer, and McCallum}]{drozdov-etal-2020-unsupervised}
Andrew Drozdov, Subendhu Rongali, Yi-Pei Chen, Tim O{'}Gorman, Mohit Iyyer, and Andrew McCallum. 2020.
\newblock \href {https://doi.org/10.18653/v1/2020.emnlp-main.392} {Unsupervised parsing with {S}-{DIORA}: Single tree encoding for deep inside-outside recursive autoencoders}.
\newblock In \emph{Proceedings of the 2020 Conference on Empirical Methods in Natural Language Processing (EMNLP)}, pages 4832--4845, Online. Association for Computational Linguistics.

\bibitem[{Drozdov et~al.(2019)Drozdov, Verga, Yadav, Iyyer, and McCallum}]{drozdov2019unsupervised}
Andrew Drozdov, Pat Verga, Mohit Yadav, Mohit Iyyer, and Andrew McCallum. 2019.
\newblock \href {http://arxiv.org/abs/1904.02142} {Unsupervised latent tree induction with deep inside-outside recursive autoencoders}.

\bibitem[{Dyer et~al.(2016)Dyer, Kuncoro, Ballesteros, and Smith}]{dyer-etal-2016-recurrent}
Chris Dyer, Adhiguna Kuncoro, Miguel Ballesteros, and Noah~A. Smith. 2016.
\newblock \href {https://doi.org/10.18653/v1/N16-1024} {Recurrent neural network grammars}.
\newblock In \emph{Proceedings of the 2016 Conference of the North {A}merican Chapter of the Association for Computational Linguistics: Human Language Technologies}, pages 199--209, San Diego, California. Association for Computational Linguistics.

\bibitem[{Hayou et~al.(2020)Hayou, Clerico, He, Deligiannidis, Doucet, and Rousseau}]{Hayou2020StableR}
Soufiane Hayou, Eugenio Clerico, Bo~He, George Deligiannidis, A.~Doucet, and J.~Rousseau. 2020.
\newblock Stable resnet.
\newblock In \emph{International Conference on Artificial Intelligence and Statistics}.

\bibitem[{Hendrycks and Gimpel(2016)}]{Hendrycks2016GaussianEL}
Dan Hendrycks and Kevin Gimpel. 2016.
\newblock \href {https://api.semanticscholar.org/CorpusID:125617073} {Gaussian error linear units (gelus)}.
\newblock \emph{arXiv: Learning}.

\bibitem[{Hong et~al.(2021)Hong, Li, Zhu, and Huang}]{Hong2021VLGrammarGG}
Yining Hong, Qing Li, Song-Chun Zhu, and Siyuan Huang. 2021.
\newblock \href {https://api.semanticscholar.org/CorpusID:232335373} {Vlgrammar: Grounded grammar induction of vision and language}.
\newblock \emph{2021 IEEE/CVF International Conference on Computer Vision (ICCV)}, pages 1645--1654.

\bibitem[{Hua et~al.(2021)Hua, Wang, Xue, Ren, Wang, and Zhao}]{9709917}
Tianyu Hua, Wenxiao Wang, Zihui Xue, Sucheng Ren, Yue Wang, and Hang Zhao. 2021.
\newblock \href {https://doi.org/10.1109/ICCV48922.2021.00946} {On feature decorrelation in self-supervised learning}.
\newblock In \emph{2021 IEEE/CVF International Conference on Computer Vision (ICCV)}, pages 9578--9588.

\bibitem[{Kim et~al.(2019{\natexlab{a}})Kim, Dyer, and Rush}]{kim2019compound}
Yoon Kim, Chris Dyer, and Alexander~M Rush. 2019{\natexlab{a}}.
\newblock Compound probabilistic context-free grammars for grammar induction.
\newblock In \emph{Proceedings of the 57th Annual Meeting of the Association for Computational Linguistics}, pages 2369--2385.

\bibitem[{Kim et~al.(2019{\natexlab{b}})Kim, Rush, Yu, Kuncoro, Dyer, and Melis}]{kim-etal-2019-unsupervised}
Yoon Kim, Alexander Rush, Lei Yu, Adhiguna Kuncoro, Chris Dyer, and G{\'a}bor Melis. 2019{\natexlab{b}}.
\newblock \href {https://doi.org/10.18653/v1/N19-1114} {Unsupervised recurrent neural network grammars}.
\newblock In \emph{Proceedings of the 2019 Conference of the North {A}merican Chapter of the Association for Computational Linguistics: Human Language Technologies, Volume 1 (Long and Short Papers)}, pages 1105--1117, Minneapolis, Minnesota. Association for Computational Linguistics.

\bibitem[{Liu et~al.(2023)Liu, Yang, Kim, and Tu}]{liu_simple_2023}
Wei Liu, Songlin Yang, Yoon Kim, and Kewei Tu. 2023.
\newblock \href {https://doi.org/10.18653/v1/2023.findings-emnlp.113} {Simple {Hardware}-{Efficient} {PCFGs} with {Independent} {Left} and {Right} {Productions}}.
\newblock In \emph{Findings of the {Association} for {Computational} {Linguistics}: {EMNLP} 2023}, pages 1662--1669, Singapore. Association for Computational Linguistics.

\bibitem[{Marcus et~al.(1994)Marcus, Kim, Marcinkiewicz, MacIntyre, Bies, Ferguson, Katz, and Schasberger}]{marcus1994penn}
Mitch Marcus, Grace Kim, Mary~Ann Marcinkiewicz, Robert MacIntyre, Ann Bies, Mark Ferguson, Karen Katz, and Britta Schasberger. 1994.
\newblock The penn treebank: Annotating predicate argument structure.
\newblock In \emph{Human Language Technology: Proceedings of a Workshop held at Plainsboro, New Jersey, March 8-11, 1994}.

\bibitem[{Papyan et~al.(2020)Papyan, Han, and Donoho}]{Papyan2020PrevalenceON}
Vardan Papyan, Xuemei Han, and David~L. Donoho. 2020.
\newblock \href {https://api.semanticscholar.org/CorpusID:221172897} {Prevalence of neural collapse during the terminal phase of deep learning training}.
\newblock \emph{Proceedings of the National Academy of Sciences of the United States of America}, 117:24652 -- 24663.

\bibitem[{Park and Kim(2024)}]{park-kim-2024-structural}
Jinwook Park and Kangil Kim. 2024.
\newblock \href {https://doi.org/10.18653/v1/2024.findings-acl.898} {Structural optimization ambiguity and simplicity bias in unsupervised neural grammar induction}.
\newblock In \emph{Findings of the Association for Computational Linguistics: ACL 2024}, pages 15124--15139, Bangkok, Thailand. Association for Computational Linguistics.

\bibitem[{Paszke et~al.(2019)Paszke, Gross, Massa, Lerer, Bradbury, Chanan, Killeen, Lin, Gimelshein, Antiga et~al.}]{paszke2019pytorch}
Adam Paszke, Sam Gross, Francisco Massa, Adam Lerer, James Bradbury, Gregory Chanan, Trevor Killeen, Zeming Lin, Natalia Gimelshein, Luca Antiga, et~al. 2019.
\newblock Pytorch: An imperative style, high-performance deep learning library.
\newblock \emph{Advances in neural information processing systems}, 32.

\bibitem[{Petrov et~al.(2006)Petrov, Barrett, Thibaux, and Klein}]{petrov2006learning}
Slav Petrov, Leon Barrett, Romain Thibaux, and Dan Klein. 2006.
\newblock Learning accurate, compact, and interpretable tree annotation.
\newblock In \emph{Proceedings of the 21st International Conference on Computational Linguistics and 44th Annual Meeting of the Association for Computational Linguistics}, pages 433--440.

\bibitem[{Seddah et~al.(2014)Seddah, K{\"u}bler, and Tsarfaty}]{seddah-etal-2014-introducing}
Djam{\'e} Seddah, Sandra K{\"u}bler, and Reut Tsarfaty. 2014.
\newblock \href {https://aclanthology.org/W14-6111} {Introducing the {SPMRL} 2014 shared task on parsing morphologically-rich languages}.
\newblock In \emph{Proceedings of the First Joint Workshop on Statistical Parsing of Morphologically Rich Languages and Syntactic Analysis of Non-Canonical Languages}, pages 103--109, Dublin, Ireland. Dublin City University.

\bibitem[{Shen et~al.(2019)Shen, Tan, Sordoni, and Courville}]{shen2019ordered}
Yikang Shen, Shawn Tan, Alessandro Sordoni, and Aaron Courville. 2019.
\newblock \href {http://arxiv.org/abs/1810.09536} {Ordered neurons: Integrating tree structures into recurrent neural networks}.

\bibitem[{Shen et~al.(2021)Shen, Tay, Zheng, Bahri, Metzler, and Courville}]{shen2021structformer}
Yikang Shen, Yi~Tay, Che Zheng, Dara Bahri, Donald Metzler, and Aaron Courville. 2021.
\newblock Structformer: Joint unsupervised induction of dependency and constituency structure from masked language modeling.
\newblock In \emph{Proceedings of the 59th Annual Meeting of the Association for Computational Linguistics and the 11th International Joint Conference on Natural Language Processing (Volume 1: Long Papers)}, pages 7196--7209.

\bibitem[{Stolcke and Omohundro(1994)}]{Stolcke_1994}
Andreas Stolcke and Stephen Omohundro. 1994.
\newblock \href {https://doi.org/10.1007/3-540-58473-0_141} {\emph{Inducing probabilistic grammars by Bayesian model merging}}, page 106–118. Springer Berlin Heidelberg.

\bibitem[{Wang et~al.(2019)Wang, Lee, and Chen}]{wang2019tree}
Yau-Shian Wang, Hung-Yi Lee, and Yun-Nung Chen. 2019.
\newblock \href {http://arxiv.org/abs/1909.06639} {Tree transformer: Integrating tree structures into self-attention}.

\bibitem[{Williams(2023)}]{Williams2023StructuredGM}
Christopher K.~I. Williams. 2023.
\newblock \href {https://api.semanticscholar.org/CorpusID:256627452} {Structured generative models for scene understanding}.
\newblock \emph{Int. J. Comput. Vis.}, 133:2845--2867.

\bibitem[{Xue et~al.(2005)Xue, Xia, Chiou, and Palmer}]{xue2005penn}
Nianwen Xue, Fei Xia, Fu-Dong Chiou, and Martha Palmer. 2005.
\newblock \href {https://doi.org/10.1017/S135132490400364X} {The penn chinese treebank: Phrase structure annotation of a large corpus}.
\newblock \emph{Natural Language Engineering}, 11:207--238.

\bibitem[{Yang et~al.(2022)Yang, Liu, and Tu}]{yang2022dynamic}
Songlin Yang, Wei Liu, and Kewei Tu. 2022.
\newblock \href {https://doi.org/10.18653/v1/2022.naacl-main.353} {Dynamic programming in rank space: Scaling structured inference with low-rank {HMM}s and {PCFG}s}.
\newblock In \emph{Proceedings of the 2022 Conference of the North American Chapter of the Association for Computational Linguistics: Human Language Technologies}, pages 4797--4809, Seattle, United States. Association for Computational Linguistics.

\bibitem[{Yang et~al.(2021{\natexlab{a}})Yang, Zhao, and Tu}]{yang2021neural}
Songlin Yang, Yanpeng Zhao, and Kewei Tu. 2021{\natexlab{a}}.
\newblock \href {https://doi.org/10.18653/v1/2021.acl-long.209} {Neural bi-lexicalized {PCFG} induction}.
\newblock In \emph{Proceedings of the 59th Annual Meeting of the Association for Computational Linguistics and the 11th International Joint Conference on Natural Language Processing (Volume 1: Long Papers)}, pages 2688--2699, Online. Association for Computational Linguistics.

\bibitem[{Yang et~al.(2021{\natexlab{b}})Yang, Zhao, and Tu}]{yang2021pcfgs}
Songlin Yang, Yanpeng Zhao, and Kewei Tu. 2021{\natexlab{b}}.
\newblock \href {https://doi.org/10.18653/v1/2021.naacl-main.117} {{PCFG}s can do better: Inducing probabilistic context-free grammars with many symbols}.
\newblock In \emph{Proceedings of the 2021 Conference of the North American Chapter of the Association for Computational Linguistics: Human Language Technologies}, pages 1487--1498, Online. Association for Computational Linguistics.

\bibitem[{Zhang and Sennrich(2019)}]{Zhang2019RootMS}
Biao Zhang and Rico Sennrich. 2019.
\newblock \href {https://api.semanticscholar.org/CorpusID:113405151} {Root mean square layer normalization}.
\newblock \emph{ArXiv}, abs/1910.07467.

\bibitem[{Zhang et~al.(2019)Zhang, Dauphin, and Ma}]{Zhang2019FixupIR}
Hongyi Zhang, Yann Dauphin, and Tengyu Ma. 2019.
\newblock Fixup initialization: Residual learning without normalization.
\newblock In \emph{International Conference on Learning Representations}.

\bibitem[{Zhao and Titov(2020)}]{Zhao2020VisuallyGC}
Yanpeng Zhao and Ivan Titov. 2020.
\newblock \href {https://api.semanticscholar.org/CorpusID:221971028} {Visually grounded compound pcfgs}.
\newblock In \emph{Conference on Empirical Methods in Natural Language Processing}.

\end{thebibliography}

\appendix

\section{Appendices}

% Experiments에서 참고됨.
\subsection{Experiment Details}
\label{appendix:exp_details}

\paragraph{Dataset Detail}
We adopt the standard setup and preprocessing for PTB,\footnote{The license of PTB is LDC User Agreement for Non-Members. \url{https://catalog.ldc.upenn.edu/LDC99T42}} using sections 02–21 for training, section 22 for validation, and section 23 for testing, with punctuation and trivial constituents removed. The vocabulary consists of the 10,000 most frequent words, while all other words are replaced with \verb|<unk>|. For data processing, we follow the pipeline employed by prior models that utilize base models and parsers, as described in~\cite{kim2019compound,yang2022dynamic,shen2021structformer,yang2021neural,drozdov2019unsupervised}.
We also utilize the CTB\footnote{The license of CTB is LDC User Agreement for Non-Members. \url{https://catalog.ldc.upenn.edu/LDC2005T01}} and the Basque, French, German, Hebrew, Hungarian, Korean, Polish, and Swedish datasets from SPMRL,\footnote{The license of SPMRL is Creative Commons Attribution 4.0 International License. \url{https://www.spmrl.org/spmrl2013-sharedtask.html}} following the standard setup used in prior work.

\paragraph{Implementation Detail}
To implement our methodology on top of the base model FGG-TNPCFGs, we utilized PyTorch version 2.2~\cite{paszke2019pytorch}. For smooth processing and analysis of tree structures, we employed NLTK\footnote{https://www.nltk.org/}.

\begin{figure*}[!ht]
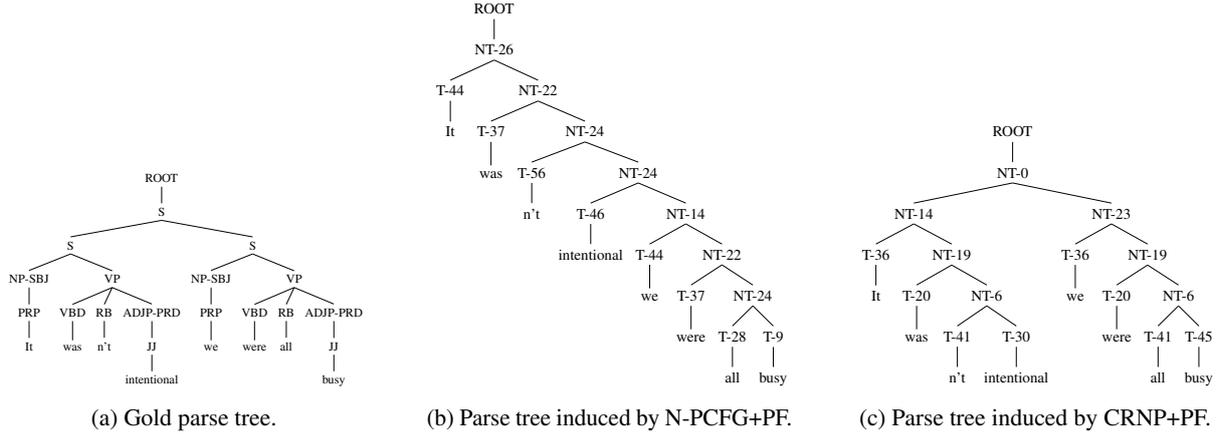

\centering
    \begin{subfigure}[b]{0.3\textwidth}
        \resizebox{\textwidth}{!}{
        \Tree [.ROOT
        [.S
          [.S
            [.NP-SBJ [.PRP It ] ]
            [.VP
              [.VBD was ]
              [.RB n't ]
              [.ADJP-PRD [.JJ intentional ] ]
            ]]
          [.S
            [.NP-SBJ [.PRP we ] ]
            [.VP
              [.VBD were ]
              [.RB all ]
              [.ADJP-PRD [.JJ busy ] ]
            ] ]
        ] ]
        }
        \caption{Gold parse tree.}
    \end{subfigure}
    \hfill
    \begin{subfigure}[b]{0.3\textwidth}
        \resizebox{\textwidth}{!}{
        \Tree [.ROOT
        [.NT-26
          [.T-44 It ]
          [.NT-22
            [.T-37 was ]
            [.NT-24
              [.T-56 n't ]
              [.NT-24
                [.T-46 intentional ]
                [.NT-14
                  [.T-44 we ]
                  [.NT-22
                    [.T-37 were ]
                    [.NT-24 [.T-28 all ] [.T-9 busy ] ]
                  ]
                ]
              ]
            ]
          ]
        ] ]
        }
        \caption{Parse tree induced by N-PCFG+PF.}
    \end{subfigure}
    \hfill
    \begin{subfigure}[b]{0.3\textwidth}
        \resizebox{\textwidth}{!}{
        \Tree [.ROOT
        [.NT-0
          [.NT-14
            [.T-36 It ]
            [.NT-19
              [.T-20 was ]
              [.NT-6
                [.T-41 n't ]
                [.T-30 intentional ]
              ]
            ]
          ]
          [.NT-23
            [.T-36 we ]
            [.NT-19
              [.T-20 were ]
              [.NT-6 [.T-41 all ] [.T-45 busy ] ]
            ]
          ]
        ] ]
        }
        \caption{Parse tree induced by CRNP+PF.}
    \end{subfigure}
\caption{Comparison between the gold parse tree and the parse trees induced by the models (N-PCFG+PF and CRNP+PF) on the example sentence \textit{"It wasn't intentional, we were all busy"}.}
\label{fig:qa_tree_app}
\end{figure*}

\paragraph{Training Detail}
The hyperparameter settings follow those reported in \citet{yang2022dynamic}. The ratio of nonterminal to preterminal symbols is primarily fixed at 1:2. To analyze the effect of the number of symbols, we conduct experiments varying the number of nonterminal symbols from 1 to 4500. Training proceeds for up to 10 epochs, with early stopping based on validation likelihood. Optimization is performed using the Adam optimizer with a learning rate of $2 \times 10^{-3}$, $\beta_1 = 0.75$, and $\beta_2 = 0.999$. The model is configured with a rank size of 1000, a symbol embedding size of 256, and a word embedding size of 200. These hyperparameters are held constant across all languages without task-specific tuning. All experiments are run on an NVIDIA RTX 2080Ti. To measure variance, we repeat each experiment 32 times for both the original FGG-TNPCFGs and our method, as reported in Table~\ref{tab:final-performance}, while all other experiments are repeated four times. Each run takes approximately 3 hours.

\subsection{Measures for probability distribution collapse}
\label{sec:measures}

\paragraph{Local Perplexity}
represents the number of valid rules in single probability distribution. In other words, local perplexity (PPL) represent sparsity of single probability distribution. Therefore, probability distribution that have low local PPL have few utilized rules in distribution and is sparse.

\begin{equation*}
    \text{PPL}_{\text{local}}=\frac{\sum_{p\in P}\exp H(p)}{|P|}
\end{equation*}

\paragraph{Global PPL}
represent the number of valid categories of the mean distribution of whole probability distributions in grammar, which represent how many various rules are used without duplication for parsing. This is related with the expressive power of grammar. In other words, higher performance, higher variety of the rules that utilized in grammar.

\begin{gather*}
    q(C)=\frac{\sum_{P'\in S}p(P'\rightarrow C)}{|S|}\\
    \text{PPL}_{\text{global}}=\exp H(q)
\end{gather*}

\paragraph{Geometric Mean of Pairwise Jensen-Shannon Divergence} shows how different the probability distributions in grammar with each other. In this paper, we use JSD with base-e logarithm for computational convenience. JSD is represented by following equation:
\begin{gather*}
    \text{JSD}(P\|Q)=\frac{1}{2}\text{KL}(P\|M)+\frac{1}{2}\text{KL}(Q\|M)\\
    \text{where }M=\frac{1}{2}(P+Q)
\end{gather*}

We use JSD to compare similarity for several distributions, then we get the values for $\frac{N(N-1)}{2}$ pairs of distributions, where $N$ represent the number of distributions.
We use geometric mean to average on these values of pairs.
The reason why we use geometric mean is arithmetic mean is not sensitive on outlier, which make that the duplication of pairs is not appeared in values in arithmetic mean.
Therefore, the GPJ is defined as:

\begin{equation*}
    \text{GPJ}=\sqrt[|\text{pairs}|]{\prod_{P,Q\in S}JSD(P\|Q)}.
\end{equation*}

\subsection{Additional Examples for parse trees}
\label{appendix:trees}

We provide additional examples of parse trees to compare the quality by methods, N-PCFG+PF and CRNP+PF in Figure~\ref{fig:qa_tree_app}.

\end{document}